\documentclass[10pt, conference, compsocconf]{IEEEtran}

\usepackage{amsmath,amsfonts}
\usepackage{algorithmic}
\usepackage{algorithm}
\usepackage{array}
\usepackage[caption=false,font=normalsize,labelfont=sf,textfont=sf]{subfig}
\usepackage{textcomp}
\usepackage{stfloats}
\usepackage{url}
\usepackage{verbatim}
\usepackage{graphicx}
\usepackage{cite}

\usepackage{booktabs} 
\usepackage{xcolor} 

\ifCLASSINFOpdf
\else
\fi
\begin{document}
%
\title{Multi-Object Tracking and Segmentation with a Space-Time Memory Network}


\author{\IEEEauthorblockN{Mehdi Miah, Guillaume-Alexandre Bilodeau, Nicolas Saunier}
\IEEEauthorblockA{Polytechnique Montréal\\
Québec, Canada\\
E-mail: mehdi.miah@polymtl.ca, gabilodeau@polymtl.ca, nicolas.saunier@polymtl.ca}
}


%


\maketitle

\begin{abstract}
We propose a method for multi-object tracking and segmentation based on a novel memory-based mechanism to associate tracklets. The proposed tracker, MeNToS, addresses particularly the long-term data association problem, when objects are not observable for long time intervals. Indeed, the recently introduced HOTA metric (High Order Tracking Accuracy), which has a better alignment than the formerly established MOTA (Multiple Object Tracking Accuracy) with the human visual assessment of tracking, has shown that improvements are still needed for data association, despite the recent improvement in object detection. In MeNToS, after creating tracklets using instance segmentation and optical flow, the proposed method relies on a space-time memory network originally developed for one-shot video object segmentation to improve the association of sequence of detections (tracklets) with temporal gaps. We evaluate our tracker on KITTIMOTS and MOTSChallenge and we show the benefit of our data association strategy with the HOTA metric. Additional ablation studies demonstrate that our approach using a space-time memory network gives better and more robust long-term association than those based on a re-identification network. Our project page is at \url{www.mehdimiah.com/mentos+}.
\end{abstract}

\begin{IEEEkeywords}
tracking; MOTS; memory network; data association

\end{IEEEkeywords}

%
\IEEEpeerreviewmaketitle

\section{Introduction}\label{sec1}

Object tracking is a common task in computer vision: given a video, the objective is to detect objects (for example, all road users such as vehicles, pedestrians and cyclists) and to attribute a unique identifier to each object. This task is fundamental for several applications such as for autonomous vehicles, city traffic management and road safety analysis~\cite{zangenehpour2016Aresignalized, fu2019Investigatingsecondary}. In the last application, since crashes between road users are not always easily observable, the use of computer vision and surrogate risk estimation metrics let researchers collect more data about road safety without relying on the actual observation of incidents. From extracted user trajectories, several safety indicators are computed such as speed, acceleration, post-encroachment time or time-to-collision~\cite{beauchampStudyAutomatedShuttle2022}. Besides, the widespread presence of cameras in cities enables large-scale studies.

However, collecting trajectories from raw video is particularly hard since the trajectories of interest are those involving at least two objects at a close distance. In such situations, it is common to observe occlusions, partial or complete, provoking missed detections and also identity switches (the identifiers of two objects are inverted). Objects trajectories are then incorrect, altering or mis-attributing the computed road safety indicators. That is why developing a tracking method robust to occlusions is particularly important for road safety indicators.

The most popular paradigm for tracking multiple object is ``tracking-by-detection'': the first step, named the detection step, detects all objects of interest in all frames of the video and the second step, named the association step, aims to assign unique consistent identities to all the objects in a video. This paper mainly focuses on the second aspect. In particular, multi-object tracking and segmentation (MOTS)~\cite{voigtlaender2019MOTSMultiObject} consists in tracking several objects \textit{at the pixel level} where the objects of interest are those belonging to a \textit{category class} such as ``cars''. Until recently, MOTS was generally evaluated with measures heavily biased against the association step~\cite{luiten2020HOTAHigher, valmadre2021LocalMetrics}. Luiten et al.~\cite{luiten2020HOTAHigher} proved that some metrics such as the Multiple Object Tracking Accuracy (MOTA)~\cite{stiefelhagen2007CLEAR2006} and its variants ignore largely the association quality. Hence, they introduced the High Order Tracking Accuracy (HOTA) metric that not only balances the detection and the association step but also has a better alignment with the human visual assessment. Therefore, evaluating MOTS with the HOTA metrics (HOTA, DetA and AssA) should lead to methods that are more robust and in turn that could improve the quality of the analysis of traffic videos.

Our proposed tracker, named MeNToS (\textbf{Me}mory \textbf{N}etwork-based \textbf{T}racker \textbf{o}f \textbf{S}egments), consists in using a propagation method originally developed for one-shot video object segmentation (OSVOS) to solve an association problem for MOTS. 
OSVOS~\cite{caelles20182018DAVIS, caelles2017OneShotVideo, caelles20192019DAVIS} is a task where the segmentation masks of all objects of interest are \textit{provided at the first frame}. It cannot rely on detections since the classes of objects of interest are unknown.  Yet, we believe that its principle can be helpful for data association.

We solve the association problem hierarchically. First, given some instance segmentation masks, we associated masks between adjacent frames. As such, masks are spatially close and visually similar and therefore a method based on a low-level information (meaning colors) may be sufficient. That is why our short-term association is based on the optical flow. After this first association, we obtain continuous tracklets of various lengths. Second, we use a space-time memory (STM) network~\cite{oh2019VideoObject}, originally developed to solve OSVOS, to associate the tracklets. A STM network can be viewed as a spatio-temporal attention mechanism~\cite{vaswani2017AttentionAll, bahdanau2015NeuralMachine} able to find a correspondence between a target and a set of queries. As opposed to OSVOS where memory networks~\cite{kumar2016Askme, sukhbaatar2015EndToEndMemory} are used to \textit{propagate} masks to the next frames, here we use them to \textit{associate} tracklets that are temporally apart.

\begin{figure*}[ht]
    \centering               
    \includegraphics[scale = 0.48, trim={0 30 0 20},clip]{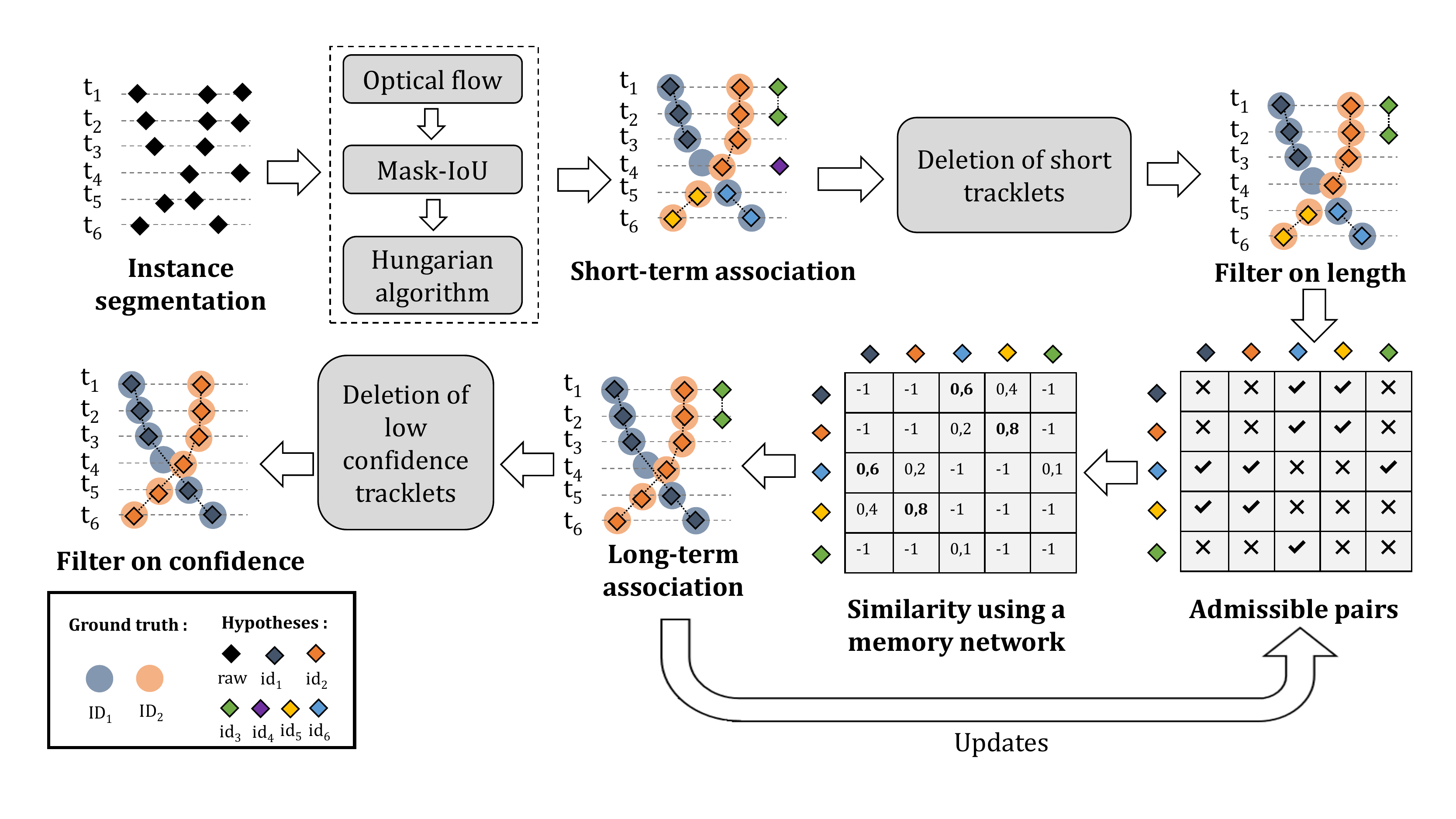}
    \caption{Illustration of our MeNToS method. Given an instance segmentation, binary masks are matched in adjacent frames to create tracklets. Very short tracklets are deleted. An appearance similarity, based on a memory network, is computed between two admissible tracklets. Then, tracklets are gradually merged starting with the pair having the highest similarity while respecting the updated constraints. Finally, low confidence tracks are deleted.}
    \label{fig:framework}
\end{figure*}

The main contributions of our work are as follows: 
\begin{itemize}
    \item We propose MeNToS, a tracker to solve MOTS based on a space-time memory network with a new similarity measure between tracklets; 
    \item We evaluate our tracker on KITTIMOTS and MOTSChallenge to prove that it is competitive, especially on the association part. We demonstrate that our approach for long-term association using a STM network is better than recent approaches based on re-identification networks;
    \item We show the usefulness of the HOTA metric for MOTS to capture the improvement resulting from improved data association.
\end{itemize}

\section{Related works}

\subsection{MOTS} 
Similarly to multi-object tracking where the ``tracking-by-detection'' paradigm is popular~\cite{dendorfer2021MOTChallengeBenchmark}, MOTS is mainly solved by creating tracklets from segmentation masks and then building long-term tracks by merging the tracklets~\cite{yang2020reMOTSSelfSupervised,zhang2020LIFTSLidar, luiten2020TrackReconstruct}. Usually, methods use an instance segmentation method to generate binary masks; ReMOTS~\cite{yang2020reMOTSSelfSupervised} used two advanced instance segmentation methods and self-supervision to refine masks. As for the association step, many methods require a re-identification (ReID) step. 
For example, Voigtlaender et al.~\cite{voigtlaender2019MOTSMultiObject} extended Mask R-CNN by an association head to return an embedding for each detection. Yang et al.~\cite{yang2020reMOTSSelfSupervised} associated two tracklets if they were temporally close, without any temporal overlap with similar appearance features based on all their observations and a hierarchical clustering. Zhang et al. ~\cite{zhang2020LIFTSLidar} used temporal attention to lower the weight of frames with occluded objects. More recently, Wei et al.~\cite{wei2021RobTrackRobust} proposed to solve MOTS by improving the quality of detected masks with massive instance augmentation during training~\cite{ghiasi2021SimpleCopyPaste} and refining masks during inference~\cite{tang2021LookCloser}, then by associating detections with an ensemble method.

\subsection{OSVOS} 
Closely related to MOTS, OSVOS requires tracking objects whose segmentation masks are only provided at the first frame. OSVOS is mainly solved by propagation-based methods: a model learns the representation of the initial mask and tries to make some correspondences in the next frames. MAST~\cite{lai2020MASTMemoryAugmented} used a memory component to predict the future location of objects. STM~\cite{oh2019VideoObject} was proposed to solve OSVOS by storing some previous frames and masks in a memory that is later read by an attention mechanism to predict the new mask in a target image. Such a network was recently used~\cite{garg2021MaskSelection} to solve video instance segmentation (VIS), a problem in which no prior knowledge is given about the objects to track. However, it is unclear how the STM network behaves when multiple instances from the same class appear in a video. We show in this work that space-time memory network performs well and can help to solve an association problem by taking advantage of the information at the pixel level and the presence of other objects.

\subsection{Bridging the gap between MOTS and OSVOS}
Despite some clear similarities between the two tasks (prediction of the future position of an object at the pixel level), there are some differences such as the evaluation (sMOTSA, HOTA measures which evaluate the quality of detection and association versus $\mathcal{J}\&\mathcal{F}$ which evaluates the masks quality with a region similarity and a contour accuracy) and the specification at the first frame (predefined classes without any initial mask versus class-agnostic with initial mask). Recently, Wang et al.~\cite{wang2021DifferentTracking} addressed this issue by considering a common shared appearance model to solve several tracking problems including MOTS and OSVOS. They compared the performance of several pre-trained vision models such as an ImageNet~\cite{russakovsky2015ImageNetLarge} pretrained architecture, MoCo~\cite{he2020MomentumContrast} or CRW~\cite{jabri2020SpaceTimeCorrespondence} architectures to get visual features. Then the authors used these representations to either propagate instances like in OSVOS or associate them like in MOTS. Moreover, for the association step, they computed a similarity score between tracklets with an attention perspective. Their method consists in the reconstruction at the patch level of all features of existing tracklets and current detections. The similarity is then the average cosine similarity between the forward and backward reconstruction. Yan et al.~\cite{yan2022GrandUnification} proposed UniTrack to solve both OSVOS and MOTS with a single framework consisting in a unified backbone for representations extraction, a unified embedding module and a unified head. In their network, the distinction between MOTS and OSVOS tasks is made with some ``target priors'' maps which provide a prior information about the objects to track. 

\subsection{Memory mechanism}

The use of a memory mechanism enables to use information from multiple frames to associate detections, specially in case of long periods of occlusion. In this case, it is highly likely that the appearance of objects has changed. Instead on relying on only the last or first frame in which an object was detected, keeping multiple appearance information coming from several frames is a logical way of improving the performance of a tracker. But it is currently unknown which frames to keep in memory : the lastest, the whole appearance information, an average of appearance or a few of them. The tracker MeMOT~\cite{cai2022MeMOTMultiObject} preserves a large spatio-temporal memory to keep the embeddings of the tracked objects over 24 frames. Conversely, Korbar and Zisserman~\cite{korbarEndtoendTrackingMultiquery2022} kept the appearance information of the first and the last five frames of each tracklet to overcome the occlusions, reducing the limitation due to the GPU memory. We show in this work that keeping in memory the appearance information from only two frames leads to competitive results.

\section{Proposed method}
\label{sec:method}

As illustrated in Figure~\ref{fig:framework}, our pipeline for tracking multiple objects is based on three main steps: detections via instance segmentation, a short-term association of segmentation masks in adjacent frames and a greedy long-term association of tracklets using a memory network.
    
\subsection{Detections}

Our method follows the ``tracking-by-detection'' paradigm. First, we use the non-overlapping instance segmentations from a pre-trained detector. Objects with a detection score higher than $\theta_d$ and a size (as the number of pixels of its binary mask) higher than $\theta_a$ are extracted. Detections with a low confidence score or with a small size are usually false positives.

\subsection{Short-term association (STA)}

During the short-term association, we associate temporally close segmentation masks between adjacent frames by computing the optical flow. Masks from the previous frames are warped and a mask IoU (mIoU) is computed between these warped masks and the masks from the next frame. As some object classes are visually similar (for example car and truck), only associating objects of the same class may lead to missing some objects due to classification errors. That is why this step is a class-agnostic association, letting the model match a car with a truck if the optical flow corresponds.

The Hungarian algorithm~\cite{kuhn1955Hungarianmethod} is used to associate masks where the cost matrix is computed based on the negative mIoU. Matched detections with a mIoU above a threshold $\theta_s$ are connected to form a tracklet and the remaining detections form new tracklets. 

Finally, the class of a tracklet is the one that accumulates the highest sum of confidence score of its detections, and tracklets with only one detection are deleted since they often correspond to false positives.

\subsection{Long-term association (LTA)}

Greedy long-term association and the use of a STM network for re-identification of tracklets are the novel contributions of our approach. Once tracklets have been created, it is necessary to link them in case of fragmentation caused, for example, by occlusion. In this long-term association, we use a space-time memory network as a similarity measure between tracklets by propagating some masks of a tracklet in the past and the future. Given a binary segmentation mask in a frame of reference and a query frame, the STM network outputs a heatmap indicating the probability of the location of the object of reference in the query frame. If a predicted heatmap of a tracklet sufficiently overlaps a mask of another tracklet, these two tracklets are linked together. Given the fact that this procedure is applied at the pixel-level on the whole image, this similarity is only computed on a selection of admissible tracklet pairs to reduce the computational cost. At this step, we point out that all tracklets have a length longer than or equal to two. 

\subsubsection{Measure of similarity between tracklets}
\label{sec:notations}

\begin{figure}[ht] \centering
    \includegraphics[scale=0.65, trim={0 0 0 0},clip]{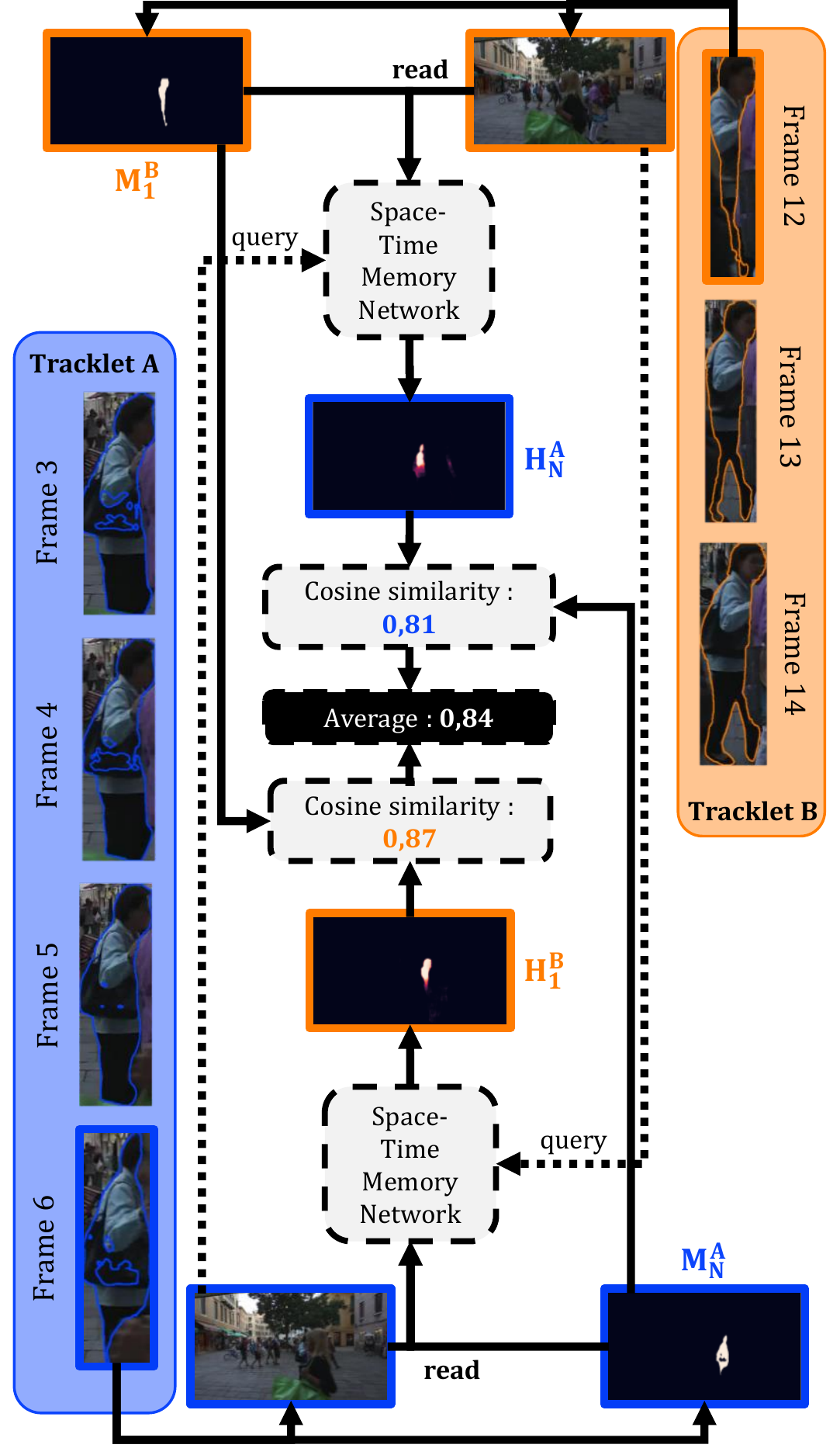}
    \caption{Similarity used at the long-term association step. For simplicity, only one mask and frame are used as reference and as target in the space-time memory network.}
    \label{fig:similarity}
\end{figure}

 Our similarity measure is based on the ability to match some parts of two different tracklets (say $T^A$ and $T^B$) and can be interpreted as a pixel-level visual-spatial alignment rather than a patch-level visual alignment~\cite{yang2020reMOTSSelfSupervised, zhang2020LIFTSLidar}. For that, we propagate some masks of tracklet $T^A$ to other frames where the tracklet $T^B$ is present and then compare the masks of $T^B$ and the propagated version of the mask heatmaps, computed before the binarization, for $T^A$. The more they are spatially aligned, the higher the similarity is. In details, let us consider two tracklets $T^A = (M_1^{A}, M_2^{A}, \cdots, M_{N}^{A})$ and $T^B = (M_1^{B}, M_2^{B}, \cdots, M_P^{B})$ of length $N$ and $P$ respectively, such that $T^A$ appears first and where $M_1^A$ denotes the first segmentation mask of the tracklet $T^A$. We use a pre-trained STM network~\cite{oh2019VideoObject} to store two binary masks as references (and their corresponding frames): the closest ones ($M_N^A$ for $T^A$ and $M_1^B$ for $T^B$) and a second mask a little farther ($M_{N-n-1}^A$ for $T^A$ and $M_n^B$ for $T^B$). The farther masks are used because the first and last object masks of a tracklet are often incomplete due, for example, to occlusions. Then, the reference frames are used as queries to produce heatmaps with continuous values between 0 and 1 ($H_N^A$, $H_{N-n-1}^A$, $H_1^B$, $H_n^B$). Finally, the average cosine similarity between these four heatmaps and the four masks ($M_N^A$, $M_{N-n-1}^A$, $M_1^B$, $M_n^B$) is the final similarity between the two tracklets, denoted as $sim(T^A, T^B)$. Figure~\ref{fig:similarity} illustrates a one-frame version of this similarity measure between tracklets. 

\subsubsection{Selection of pairs of tracklets} 

Instead of estimating a similarity measure between all pairs of tracklets, a selection is made to reduce the computational cost~\cite{yang2020reMOTSSelfSupervised}. The selection is based on the following heuristic: two tracklets may belong to the same objects if they belong to the same object class, are temporally close, spatially close and with a small temporal overlap. 

In details, let us denote $f(M)$ the frame where the mask $M$ is present, $\bar{M}$ its center and $fps, H$ and $W$ respectively the number of frames per second, height and width of the video. The temporal ($C_t(T^A, T^B)$), spatial ($C_s(T^A, T^B)$) and temporal overlap ($C_o(T^A, T^B)$) costs between $T^A$ and $T^B$ are defined respectively as: 

\begin{equation}
    C_t(T^A, T^B) = \dfrac{\vert f(M_{N}^{A}) - f(M_1^{B}) \vert}{fps},
    \label{eq:Ct}
\end{equation}
\begin{equation}
    C_s(T^A, T^B) = \dfrac{2}{H+W} \times \| \bar{M_N^A} - \bar{M_1^B} \|_1,
\label{eq:Cs}
\end{equation}
\begin{equation}
C_o(T^A, T^B) = \vert \bigcap \limits_{T \in T^A, T^B} \{f(M), \forall M \in T\} \vert
\label{eq:Co}
\end{equation}

A pair $(T^A, T^B)$ is admissible if the tracklets belong to the same object class, $C_t(T^A, T^B) \leq \tau_t$, $C_s(T^A, T^B) \leq \tau_s$ and $C_o(T^A, T^B) \leq \tau_o$.

\subsubsection{Greedy association}

 We gradually merge the admissible pairs with the highest cosine similarity, $sim(T^A, T^B)$, if it is above a threshold $\theta_l$, while continuously updating the admissible pairs using equation \ref{eq:Co}. A tracklet can therefore be repeatedly merged with other tracklets. Finally, tracks having their highest detection score lower than $\theta_f$ are deleted.

\section{Experiments}
\subsection{Implementation details}

At the detection step, the detection are provided by the RobMOTS challenge~\cite{luiten2021RobMOTSBenchmark} obtained from a Mask R-CNN X-152~\cite{he2017MaskRCNN} and Box2Seg Network~\cite{luiten2018PReMVOSProposalGeneration} for all 80 categories of COCO. The threshold of the detection confidence score $\theta_d$ is set to 50~\% and small masks whose area is less than $\theta_a=128$ pixels are removed. 

For the short-term data association, the optical flow is computed with RAFT~\cite{teed2020RAFTRecurrent}. We select $\theta_s = 0.15$ for the threshold in the Hungarian algorithm. 

For the long-term data association, the pre-selection is done with $(\tau_t, \tau_s, \tau_o) = (1.5, 0.2, 1)$. We took a space-time memory network pretrained on DAVIS~\cite{pont-tuset20182017DAVIS}. To have a more generic tracker, STM was not fine-tuned on MOTSChallenge or KITTIMOTS. To measure similarity, the second frame is picked using $n=5$. If that frame is not available, $n=2$, is used instead. We select $\theta_l = 0.30$ in the greedy association. Then, the tracklets having at least one observation with a confidence score higher than $\theta_f = 90\%$ are kept. All these hyper-parameters were selected using the HOTA score on the validation sets and \emph{remain fixed regardless of the dataset and object classes}.

\subsection{Datasets and performance evaluations}

We evaluated our method on KITTIMOTS and MOTSChallenge~\cite{voigtlaender2019MOTSMultiObject} two common datasets about MOTS. KITTIMOTS contains 21 training videos and 29 test videos on cars and pedestrians obtained from a camera mounted on the roof of a car. The scenes are captured at 10~Hz displaying real-world traffic situations. MOTSChallenge contains 4 training videos and 4 test videos only with pedestrians in cities. 
Tracking performance is measured with metrics including MOTA~\cite{stiefelhagen2007CLEAR2006} (and its variant sMOTSA, MOTSA and sMOTSP), Identity F1 score (IDF1) which measures the quality of trajectory identity accuracy, Identity switches (IDSw) which counts the number of inversion of identity. We also use, when possible, the recently introduced HOTA metric~\cite{luiten2020HOTAHigher} that fairly balances the quality of detections and associations. It can be decomposed into the DetA and AssA metrics to measure the quality of these two components. The higher the HOTA is, the more the tracker is aligned with the human visual assessment.

\subsection{Results}

Results\footnote{Full results at \url{https://www.cvlibs.net/datasets/kitti/eval_mots.php} for KITTIMOTS and \url{https://motchallenge.net/results/MOTS/} for MOTSChallenge} in tables~\ref{tab:results_kittimots} and \ref{tab:results_motschallenge} indicate that our method is competitive on the association performance. We ranked first and second respectively on pedestrians and cars on this criteria on KITTIMOTS and our identity switches are the lowest on MOTSChallenge with the second highest IDF1. \emph{Contrarily to other methods, we point out that ours does not require any additional fine-tuning on the benchmarks or per benchmark hyper-parameters selection}. For example, we did not fine-tune STM on the benchmarks. Note also that the methods in the tables do not use all the same detection inputs. The ablation studies in the next section gives a better understanding of our contribution by fixing the detection inputs to assess only the data association component.

\begin{table*}[ht] %
\begin{center}
\caption{Results on the test set of KITTIMOTS. \textbf{\textcolor{red}{bold red}} and \textit{\textcolor{blue}{italic blue}} indicate respectively the first and second best methods.}
\label{tab:results_kittimots}
\begin{tabular}{@{}llllllll@{}}
\toprule
Method & \multicolumn{3}{c}{Cars} & \multicolumn{3}{c}{Pedestrians} \\
       & {\scriptsize HOTA}$\uparrow$ & {\scriptsize DetA}$\uparrow$ & {\scriptsize AssA}$\uparrow$  & {\scriptsize HOTA}$\uparrow$ & {\scriptsize DetA}$\uparrow$ & {\scriptsize AssA}$\uparrow$ \\
\midrule
ViP-DeepLab~\cite{qiao2020ViPDeepLabLearning}   & \textbf{\textcolor{red}{76.4}} & \textbf{\textcolor{red}{82.7}} & 70.9 & \textit{\textcolor{blue}{64.3}} & \textbf{\textcolor{red}{70.7}} & \textit{\textcolor{blue}{59.5}} \\
EagerMOT~\cite{kim2020EagerMOTRealtime}         & 74.7 & 76.1 & \textit{\textcolor{blue}{73.8}} & 57.7 & 60.3 & 56.2 \\
MOTSFusion~\cite{luiten2020TrackReconstruct}    & 73.6 & 75.4 & 72.4 & 54.0 & 60.8 & 49.5 \\
ReMOTS~\cite{yang2020reMOTSSelfSupervised}      & 71.6 & 78.3 & 66.0 & 58.8 & 68.0 & 52.4 \\
PointTrack~\cite{xu2020SegmentPoints}           & 62.0 & \textit{\textcolor{blue}{79.4}} & 48.8 & 54.4 & 62.3 & 48.1 \\
\midrule
MeNToS (ours)       & \textit{\textcolor{blue}{75.8}} & 77.1 & \textbf{\textcolor{red}{74.9}} & \textbf{\textcolor{red}{65.4}} & \textit{\textcolor{blue}{68.7}} & \textbf{\textcolor{red}{63.5}} \\
\bottomrule
\end{tabular}
\end{center}
\end{table*}

\begin{table*}
\begin{center}
\caption{Results on the test set of MOTSChallenge. \textbf{\textcolor{red}{bold red}} and \textit{\textcolor{blue}{italic blue}} indicate respectively the first and second best methods.}
\label{tab:results_motschallenge}
\begin{tabular}{@{}lllllll@{}}
\toprule
Method & {\scriptsize sMOTSA}$\uparrow$& {\scriptsize IDF1}$\uparrow$ & {\scriptsize MOTSA}$\uparrow$ & {\scriptsize FP}$\downarrow$ & {\scriptsize FN}$\downarrow$ & {\scriptsize IDSw}$\downarrow$ \\
\midrule
ReMOTS~\cite{yang2020reMOTSSelfSupervised}       & \textbf{\textcolor{red}{70.4}} & \textbf{\textcolor{red}{75.0}} & \textbf{\textcolor{red}{84.4}} & \textit{\textcolor{blue}{819}} & \textit{\textcolor{blue}{3999}} & 231\\
GMPHD~\cite{song2020OnlineMultiObject}           & \textit{\textcolor{blue}{69.4}} & 66.4 & \textit{\textcolor{blue}{83.3}} & 935 & \textbf{\textcolor{red}{3985}} & 484\\
MPNTrack~\cite{braso2022MultiObjectTracking}           & 58.6 & 68.8 & 73.7 & 1059 & 7233 & \textit{\textcolor{blue}{202}}\\

SORTS~\cite{ahrnbom2021Realtimeonline}           & 55.0 & 57.3 & 68.3 & 1076 & 8598 & 552\\
TrackRCNN~\cite{voigtlaender2019MOTSMultiObject} & 40.6 & 42.4 & 55.2 & 1261 & 12641 & 567\\
\midrule
MeNToS (ours)        & 64.8 & \textit{\textcolor{blue}{73.2}} & 76.9 & \textbf{\textcolor{red}{654}} & 6704 & \textbf{\textcolor{red}{110}}\\
\bottomrule
\end{tabular}
\end{center}
\end{table*}

\begin{figure*}[ht]
    \centering               
    \includegraphics[scale = 0.44, trim={60 365 45 30},clip]{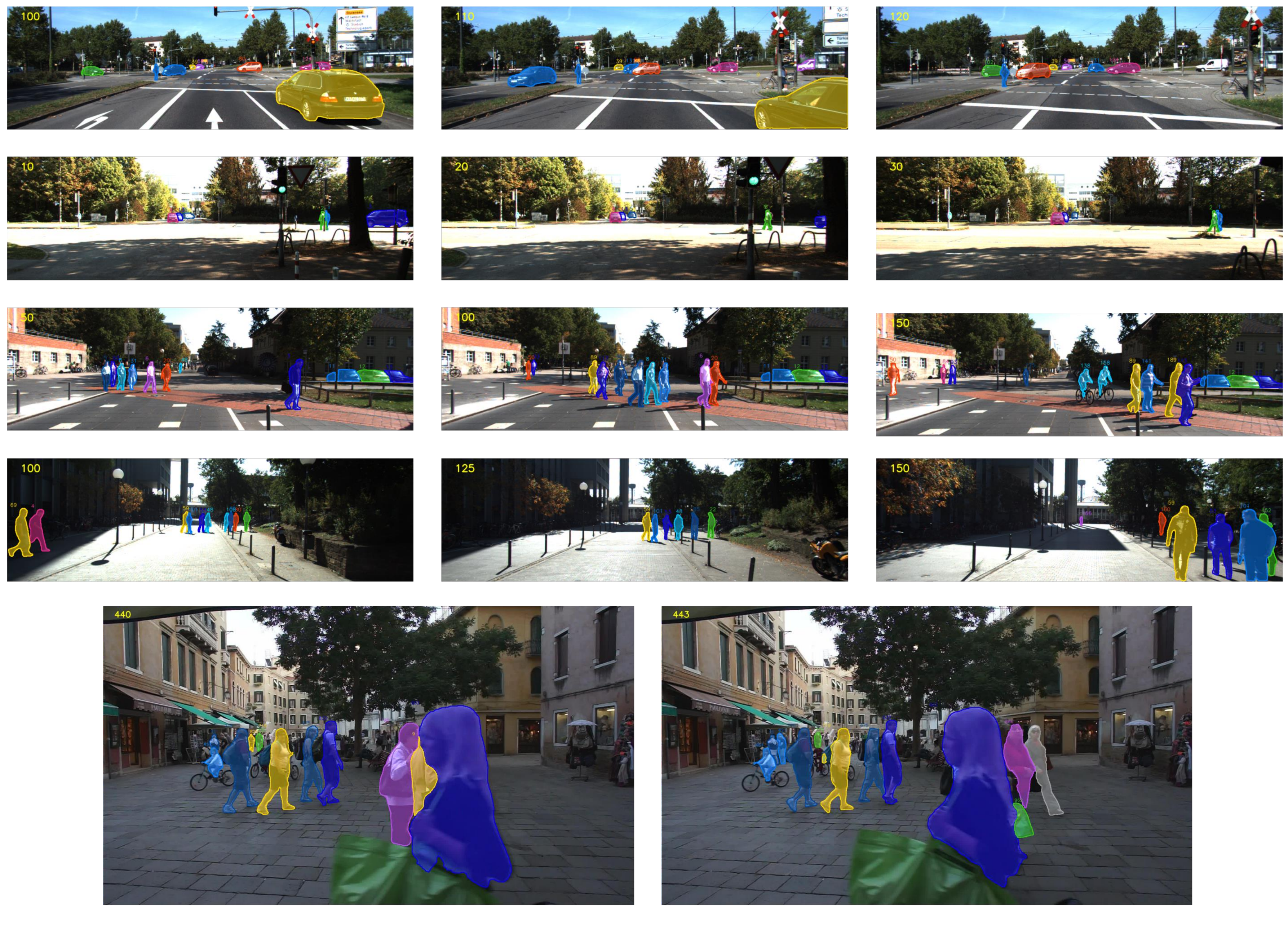}
    \caption{Qualitative results on KITTIMOTS and MOTSChallenge. Each row corresponds to a subsequence of a video clip.}
    \label{fig:qualitative_results}
\end{figure*}

Some qualitative results are provided in Figure~\ref{fig:qualitative_results}. We notice that our method can successfully retrieve objects even after occlusions, as happens to the green, yellow and purple cars in the first sequence and the blue pedestrian in the second sequence. The main limitations are contaminated segmentation masks and short tracklets since they provide little information and their appearances are unstable. For example, pedestrians usually walk in group provoking more occlusions and eventually lowering the quality of the masks, as in the third and fourth sequences. A large spatial displacement of the center of the masks may happen in the case of an occlusion, like in the fifth sequence. In that particular case, the spatial distance between the center of the two masks (the yellow one in the first frame and the green one in the second one) is higher than the threshold used for the constraint in equation~\ref{eq:Cs}. 

\section{Ablation studies}

\subsection{Contribution of each step}

To examine the effects of each step, we evaluate the performance by successively adding a component at a time. 

\begin{table}[ht]
\begin{center}
\caption{ablation studies on the validation set of KITTIMOTS (KT) and the train set of MOTSChallenge. each step of our approach leads to an improvement in terms of HOTA and sMOTSA.}
\label{tab:steps}
\begin{tabular}{p{1.1cm}p{1.1cm}p{1.1cm}p{1.1cm}p{1.1cm}}
\toprule
Step & KT-car & KT-ped & \multicolumn{2}{c}{MOTSChallenge}\\
       & HOTA & HOTA & HOTA & sMOTSA \\
\midrule
STA     & 79.4 & 62.0 & 57.9 & 62.8 \\
+ filter & 79.8 {\footnotesize (\textcolor[rgb]{0, 0.5, 0}{+0.4})} & 63.3 {\footnotesize (\textcolor[rgb]{0, 0.5, 0}{+1.3})} & 58.2 {\footnotesize (\textcolor[rgb]{0, 0.5, 0}{+0.3})} & 64.0 {\footnotesize (\textcolor[rgb]{0, 0.5, 0}{+1.2})}\\
+ LTA   & 84.1 {\footnotesize (\textcolor[rgb]{0, 0.5, 0}{+4.3})} & 66.2 {\footnotesize (\textcolor[rgb]{0, 0.5, 0}{+2.9})} & 64.5 {\footnotesize (\textcolor[rgb]{0, 0.5, 0}{+6.3})} & 64.9 {\footnotesize (\textcolor[rgb]{0, 0.5, 0}{+0.9})}\\
+ filter & 84.1 {\footnotesize (\textcolor[rgb]{0, 0.0, 0}{+0.0})} & 67.6 {\footnotesize (\textcolor[rgb]{0, 0.5, 0}{+1.4})} & 65.4 {\footnotesize (\textcolor[rgb]{0, 0.5, 0}{+0.9})} & 67.7 {\footnotesize (\textcolor[rgb]{0, 0.5, 0}{+2.8})}\\
\bottomrule
\end{tabular}
\end{center}
\end{table}

Results of Table~\ref{tab:steps} indicate that all steps improve the tracking performance (either in terms of HOTA or sMOTSA). More precisely, on KITTIMOTS and MOTSChallenge, the HOTA metric is highly sensitive to the quality of the association. From the tracking results obtained at the STA step, on average, two thirds of the total improvement in HOTA is made at the LTA step. As for sMOTSA, unsurprisingly, it shows a smaller improvement brought by the long-term association. Consequently, improving the LTA step leads to a boost in terms of HOTA: this improvement in the data association is not fully captured by the previous sMOTSA measure.

\subsection{Upper bounds with oracles}

Since the HOTA gives more importance to the data association part of MOTS compared to the sMOTSA, we also conducted some experiments to measure the current limitations of our method and estimate the steps which can lead to the biggest gain in HOTA. Using the ground truth annotations on KITTIMOTS, we build two methods that incorporate some ground truth knowledge:
\begin{enumerate}
 \item \textbf{OracleLTA} that perfectly associates tracklets (perfect long-term association). Precisely, after associating masks using the optical flow with RAFT and filtering short tracklets, we compute the ground truth identity for each tracklet and associate them based on the same ground truth identity. This oracle ignores the step of the selection of candidate pairs. Then tracklets that do not correspond to any ground truth one are deleted.

\item \textbf{OracleSLTA} that perfectly associates detections from all frames (perfect short-term and long-term association). The oracleSLTA is obtained with perfect short and long-term assignments of detections that match the ground truth and a perfect deletion of false positives. Precisely, this is the highest upper bound using the ground truth knowledge after the detection step. For all segmentation masks, if they correspond to a ground truth mask with a mIoU higher than 50\%, they are kept and the ground truth identity is attributed, otherwise they are removed.
\end{enumerate}

We evaluated the HOTA of these two oracle upper bounds on the validation set of KITTIMOTS and compare them with our method. Results from Table~\ref{tab:oracle} indicate that MeNToS is able to reach performance close to the oracles, although improvements are still possible. Potential gains of 3.3 and 5.1 points are still possible on the car and pedestrian classes of KITTIMOTS. Moreover, the short-term association step with RAFT is nearly perfect for cars: compared to the tracker with perfect association at both short and long-term association, 70\% of the gap comes from the long-term association step. On the contrary, the short-term association is the limiting step for pedestrian: MeNToS is close to the performance of the tracker using oracle information at the long-term association step. We hypothesize that the difference between the car and pedestrian classes comes from the fact that there are more distractors (detected objects wrongly classified, such as mannequins recognized as persons), for pedestrians than for cars. Indeed, the model fully using the ground truth annotations is able to eliminate all distractors. Moreover, pedestrians are harder to track in KITTIMOTS because of their deformable nature and of their vertical shape which moves more on the horizontal axis between frames when the car capturing the videos is moving.

\begin{figure*}[ht] \centering
    \includegraphics[scale=0.48, trim={110 0 0 0},clip]{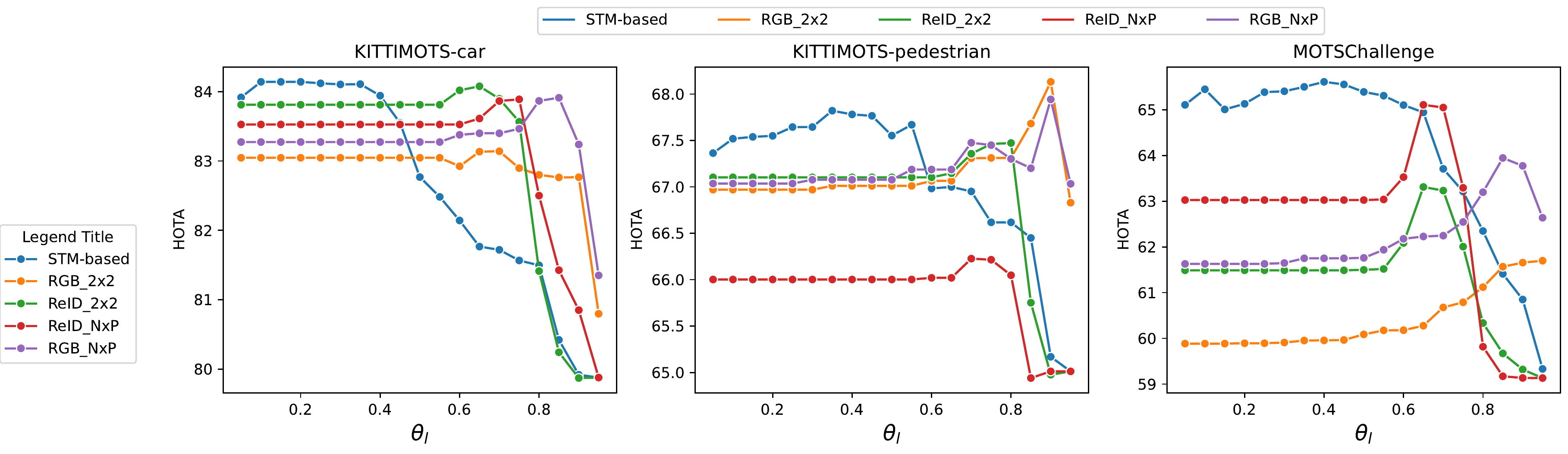}
    \caption{Comparison of some strategies of the long-term association for KITTIMOTS-car, KITTIMOTS-pedestrian and MOTSChallenge.}
    \label{fig:comparison_methodSIM}
\end{figure*}

\begin{table}[ht] %
\begin{center}
\caption{HOTA for the oracle methods on the validation set of KITTIMOTS (KT).}
\label{tab:oracle}
\begin{tabular}{@{}lcccc@{}}
\toprule
& & & \multicolumn{2}{c}{Use of oracle} \\
Method & KT-car & KT-ped & at STA & at LTA \\
\midrule
MeNToS      & 84.1 & 67.6 & - & - \\
OracleLTA   & 86.4 (\textcolor[rgb]{0, 0.5, 0}{+2.3}) & 69.4 (\textcolor[rgb]{0, 0.5, 0}{+1.8}) & - & \checkmark \\
OracleSLTA  & 87.4 (\textcolor[rgb]{0, 0.5, 0}{+3.3}) & 72.7 (\textcolor[rgb]{0, 0.5, 0}{+5.1}) & \checkmark & \checkmark \\
\bottomrule
\end{tabular}
\end{center}
\end{table}

\subsection{Comparison with other strategies of long-term association}

In the proposed method, we use a STM network to compute a similarity measure between tracklets. Other methods traditionally used re-identification features~\cite{yang2020reMOTSSelfSupervised} or color histograms~\cite{xing2009Multiobjecttracking}. Moreover, they also tend to use more reference frames whereas ours only takes into account two frames. In order to measure the contribution of the STM network combined with a cosine similarity, we replace them by one the following method: 
\begin{enumerate}
    \item \texttt{RGB\_2x2}: for two tracklets, we extracted the histogram of color of the same two frames of references computed at the pixel level. The similarity measure is then the average of four Bhattacharyya coefficients; 
    \item \texttt{RGB\_NxP}: for two tracklets, we extracted the histogram of color of all frames of reference computed on the mask. The similarity measure is then the average Bhattacharyya coefficient over all possible combinations of histograms between the two tracklets;
    \item \texttt{ReID\_2x2}: for two tracklets, we computed the ReID features of the same two frames of reference. Their similarity is the average of four cosine similarities;
    \item \texttt{ReID\_NxP}: for two tracklets, we computed the ReID features of all frames of reference. Their similarity is the average cosine similarity over all possible combinations of ReID features between the two tracklets.
\end{enumerate}
The ReID features are computed with the OsNet-AIN~\cite{zhou2019OmniScaleFeature} network. This is motivated by some works~\cite{miah2021EmpiricalAnalysis, wojke2017Simpleonline, ristani2018FeaturesMultiTarget} which indicate that ReID features are suitable for associating far apart detections and that color histograms are efficient specially for small objects.

Figure~\ref{fig:comparison_methodSIM} illustrates the behavior of these four methods for LTA alongside the one based on the STM network on HOTA with regard to the similarity threshold $\theta_l$. First, our STM-based approach is the one with the highest HOTA and is less sensitive to the parameter $\theta_l$: the HOTA is the highest on a large interval ($[0.1 ; 0.5]$). It is generally better than other methods requiring less frames in memory. Second, the four additional methods are sensitive to the parameter $\theta_l$: the interval where the HOTA is the highest is narrow and there is no single method that is better on all datasets. Then, for color histogram-based methods, using all frames of a tracklet as reference leads to a higher HOTA than using only two frames of reference. 

Generally STM-based association performs better, but it can face some issues when different parts of an object are occluded. For example, if one tracklet displays only the hood of a car and another only the car trunk, even if they belong to the same object, our STM-based approach could not associate such tracklets since no particular parts are in common. As for ReID features, they are trained to overlook such events but in the case of heavy occlusion, the features would also probably be too dissimilar. To reduce this effect, we performed the propagation at the pixel level and on the whole image for better precision to match tracklets. Working on the entire image is advantageous when a detection is missing. Moreover, this is the closest form of use of a STM network within the framework of the OSVOS.

\subsection{Number of frames of reference}

From the results of Figure~\ref{fig:comparison_methodSIM}, it seems that taking into account more reference frames leads to better results. Is this trend still relevant for our method based on STM? Since it is not technically possible with our GPU to load all frames in memory and compute the attention, we compared the following approaches: 
\begin{enumerate}
    \item \texttt{Frame 1}: only the closest frames ($f(M_N^A)$ and $f(M_1^B)$ with the notation used in section~\ref{sec:notations}) are used as references, as illustrated in Figure~\ref{fig:similarity}; 
    \item \texttt{Frames 1-2}: the two closest frames ($f(M_N^A)$, $f(M_{N-1}^A)$ and $f(M_1^B)$, $f(M_2^B)$) are used at reference; 
    \item \texttt{Frames 1-5/2}: this is the method currently used in our approach. The closest frames are used and $f(M_{N-4}^A)$ (respectively $f(M_5^B)$) if it exists for $T^A$ (respectively $T^B$), otherwise $f(M_{N-1}^A)$ (respectively $f(M_2^B)$);
    \item \texttt{Frames 1-2-5}: the two closest frames are used and $f(M_{N-4}^A)$ (respectively $f(M_5^B)$) if it exists for $T^A$ (respectively $T^B$).
\end{enumerate}

\begin{figure}[ht] %
    \centering               
    \includegraphics[scale = 0.51, trim={0 0 0 0},clip]{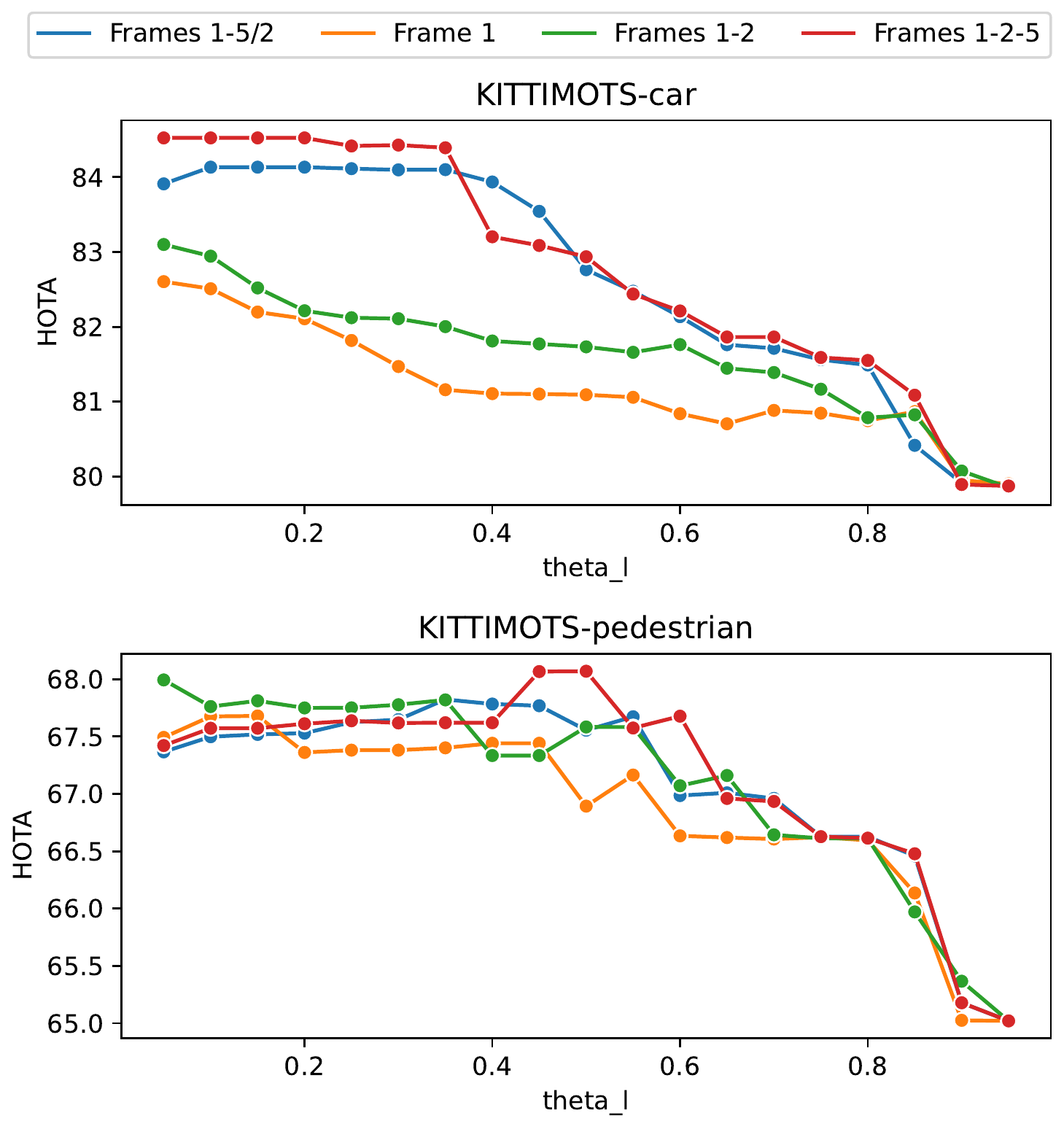}
    \caption{Ablation studies on the selection of the reference frames in the STM-based method on the validation set of KITTIMOTS.}
    \label{fig:numberFrames}
\end{figure}

Figure~\ref{fig:numberFrames} illustrates the performance in terms of HOTA for different selections of reference frames in the STM with regard to the similarity threshold $\theta_l$. On KITTIMOTS, using the closest frames as reference leads to the lowest HOTA. For pedestrians, using at least two reference frames provides a little improvement. With more frames, it seems that the performance is saturating. We hypothesize that this is due to some masks contaminated by other pedestrians, as people tend to walk in groups. As for cars, replacing the second closest frames by its fifth equivalent contributes to a boost in terms of HOTA. Considering three frames seems to saturate the HOTA. We hypothesize that using the fifth frame as a reference is a better choice than the second one because it is less similar to the first frame. Hence, more uncorrelated information is available for the STM. These results are similar to the conclusion drawn by Lai~et~al~\cite{lai2020MASTMemoryAugmented} on OSVOS: their Memory-Augmented Tracker used both short and long term memory to recover objects.



\section{Conclusion}

We propose a method to solve MOTS using a space-time memory network originally developed for solving OSVOS. It is mainly based on a hierarchical association where masks are first associated between adjacent frames to form tracklets. Then, these tracklets are associated using a STM network, leveraging the ability of the network to match similar parts. Experiments on KITTIMOTS and MOTSChallenge show that our approach gives good performance with regard to the association quality. We demonstrate that our approach for long-term association using a STM network is better than the recent approach based on re-identification networks and is less sensitive to hyper-parameters. 

\section*{Acknowledgment}
We acknowledge the support of the Natural Sciences and Engineering Research Council of Canada (NSERC), [DGDND-2020-04633 and RGPIN-2017-06115]. 

\bibliographystyle{IEEEtran}
\bibliography{references}

\begin{thebibliography}{10}
\providecommand{\url}[1]{#1}
\csname url@samestyle\endcsname
\providecommand{\newblock}{\relax}
\providecommand{\bibinfo}[2]{#2}
\providecommand{\BIBentrySTDinterwordspacing}{\spaceskip=0pt\relax}
\providecommand{\BIBentryALTinterwordstretchfactor}{4}
\providecommand{\BIBentryALTinterwordspacing}{\spaceskip=\fontdimen2\font plus
\BIBentryALTinterwordstretchfactor\fontdimen3\font minus
  \fontdimen4\font\relax}
\providecommand{\BIBforeignlanguage}[2]{{%
\expandafter\ifx\csname l@#1\endcsname\relax
\typeout{** WARNING: IEEEtran.bst: No hyphenation pattern has been}%
\typeout{** loaded for the language `#1'. Using the pattern for}%
\typeout{** the default language instead.}%
\else
\language=\csname l@#1\endcsname
\fi
#2}}
\providecommand{\BIBdecl}{\relax}
\BIBdecl

\bibitem{zangenehpour2016Aresignalized}
S.~Zangenehpour, J.~Strauss, L.~F. Miranda-Moreno, and N.~Saunier,
  ``\BIBforeignlanguage{en}{Are signalized intersections with cycle tracks
  safer? {A} case–control study based on automated surrogate safety analysis
  using video data},'' \emph{\BIBforeignlanguage{en}{Accident Analysis \&
  Prevention}}, vol.~86, pp. 161--172, Jan. 2016.

\bibitem{fu2019Investigatingsecondary}
T.~Fu, W.~Hu, L.~Miranda-Moreno, and N.~Saunier,
  ``\BIBforeignlanguage{en}{Investigating secondary pedestrian-vehicle
  interactions at non-signalized intersections using vision-based trajectory
  data},'' \emph{\BIBforeignlanguage{en}{Transportation Research Part C:
  Emerging Technologies}}, vol. 105, pp. 222--240, Aug. 2019.

\bibitem{beauchampStudyAutomatedShuttle2022}
E.~Beauchamp, N.~Saunier, and M.-S. Cloutier, ``\BIBforeignlanguage{en}{Study
  of automated shuttle interactions in city traffic using surrogate measures of
  safety},'' \emph{\BIBforeignlanguage{en}{Transportation Research Part C:
  Emerging Technologies}}, vol. 135, p. 103465, Feb. 2022.

\bibitem{voigtlaender2019MOTSMultiObject}
P.~Voigtlaender, M.~Krause, A.~Osep, J.~Luiten, B.~B.~G. Sekar, A.~Geiger, and
  B.~Leibe, ``{MOTS}: {Multi}-{Object} {Tracking} and {Segmentation},'' in
  \emph{{CVPR}}, 2019.

\bibitem{luiten2020HOTAHigher}
J.~Luiten, A.~Osep, P.~Dendorfer, P.~Torr, A.~Geiger, L.~Leal-Taixe, and
  B.~Leibe, ``{HOTA}: {A} {Higher} {Order} {Metric} for {Evaluating}
  {Multi}-{Object} {Tracking},'' \emph{International Journal of Computer Vision
  (IJCV)}, Oct. 2020.

\bibitem{valmadre2021LocalMetrics}
J.~Valmadre, A.~Bewley, J.~Huang, C.~Sun, C.~Sminchisescu, and C.~Schmid,
  ``Local {Metrics} for {Multi}-{Object} {Tracking},'' \emph{arXiv:2104.02631},
  Apr. 2021.

\bibitem{stiefelhagen2007CLEAR2006}
R.~Stiefelhagen, K.~Bernardin, R.~Bowers, J.~Garofolo, D.~Mostefa, and
  P.~Soundararajan, ``\BIBforeignlanguage{en}{The {CLEAR} 2006 {Evaluation}},''
  in \emph{\BIBforeignlanguage{en}{Multimodal {Technologies} for {Perception}
  of {Humans}}}, ser. Lecture {Notes} in {Computer} {Science}.\hskip 1em plus
  0.5em minus 0.4em\relax Berlin, Heidelberg: Springer, 2007, pp. 1--44.

\bibitem{caelles20182018DAVIS}
S.~Caelles, A.~Montes, K.-K. Maninis, Y.~Chen, L.~Van~Gool, F.~Perazzi, and
  J.~Pont-Tuset, ``The 2018 {DAVIS} {Challenge} on {Video} {Object}
  {Segmentation},'' \emph{arXiv:1803.00557}, Mar. 2018.

\bibitem{caelles2017OneShotVideo}
S.~Caelles, K.-K. Maninis, J.~Pont-Tuset, L.~Leal-Taixe, D.~Cremers, and
  L.~Van~Gool, ``One-{Shot} {Video} {Object} {Segmentation},'' in
  \emph{{CVPR}}, 2017.

\bibitem{caelles20192019DAVIS}
S.~Caelles, J.~Pont-Tuset, F.~Perazzi, A.~Montes, K.-K. Maninis, and
  L.~Van~Gool, ``The 2019 {DAVIS} {Challenge} on {VOS}: {Unsupervised}
  {Multi}-{Object} {Segmentation},'' \emph{arXiv:1905.00737}, May 2019.

\bibitem{oh2019VideoObject}
S.~W. Oh, J.-Y. Lee, N.~Xu, and S.~J. Kim, ``Video {Object} {Segmentation}
  {Using} {Space}-{Time} {Memory} {Networks},'' in \emph{{ICCV}}, 2019.

\bibitem{vaswani2017AttentionAll}
A.~Vaswani, N.~Shazeer, N.~Parmar, J.~Uszkoreit, L.~Jones, A.~N. Gomez,
  L.~Kaiser, and I.~Polosukhin, ``Attention is {All} you {Need},'' in
  \emph{{NIPS}}, 2017.

\bibitem{bahdanau2015NeuralMachine}
D.~Bahdanau, K.~Cho, and Y.~Bengio, ``Neural {Machine} {Translation} by
  {Jointly} {Learning} to {Align} and {Translate},'' in \emph{{ICLR}}, 2015.

\bibitem{kumar2016Askme}
A.~Kumar, O.~Irsoy, P.~Ondruska, M.~Iyyer, J.~Bradbury, I.~Gulrajani, V.~Zhong,
  R.~Paulus, and R.~Socher, ``Ask me anything: {Dynamic} memory networks for
  natural language processing,'' in \emph{{ICML}}, 2016.

\bibitem{sukhbaatar2015EndToEndMemory}
S.~Sukhbaatar, A.~Szlam, J.~Weston, and R.~Fergus, ``End-{To}-{End} {Memory}
  {Networks},'' in \emph{{NIPS}}, 2015.

\bibitem{dendorfer2021MOTChallengeBenchmark}
P.~Dendorfer, A.~Osep, A.~Milan, K.~Schindler, D.~Cremers, I.~Reid, S.~Roth,
  and L.~Leal-Taixé, ``\BIBforeignlanguage{en}{{MOTChallenge}: {A} {Benchmark}
  for {Single}-{Camera} {Multiple} {Target} {Tracking}},''
  \emph{\BIBforeignlanguage{en}{International Journal of Computer Vision
  (IJCV)}}, vol. 129, no.~4, pp. 845--881, Apr. 2021.

\bibitem{yang2020reMOTSSelfSupervised}
F.~Yang, X.~Chang, C.~Dang, Z.~Zheng, S.~Sakti, S.~Nakamura, and Y.~Wu,
  ``{ReMOTS}: {Self}-{Supervised} {Refining} {Multi}-{Object} {Tracking} and
  {Segmentation},'' in \emph{{CVPR} - {Workshops}}, 2020.

\bibitem{zhang2020LIFTSLidar}
H.~Zhang, Y.~Wang, J.~Cai, H.-M. Hsu, H.~Ji, and J.-N. Hwang, ``{LIFTS}:
  {Lidar} and monocular image fusion for multi-object tracking and
  segmentation,'' in \emph{{CVPR} - {Workshops}}, 2020.

\bibitem{luiten2020TrackReconstruct}
J.~Luiten, T.~Fischer, and B.~Leibe, ``Track to {Reconstruct} and {Reconstruct}
  to {Track},'' \emph{IEEE Robotics and Automation Letters}, vol.~5, no.~2, pp.
  1803--1810, Apr. 2020.

\bibitem{wei2021RobTrackRobust}
D.~Wei, J.~Hua, H.~Wang, B.~Lai, K.~Huang, C.~Zhou, J.~Huang, and X.~Hua,
  ``{RobTrack} : {A} {Robust} {Tracker} {Baseline} towards {Real}-{World}
  {Robustness} in {Multi}-{Object} {Tracking} and {Segmentation},'' in
  \emph{{CVPR} {RVSU} {Workshop}}, 2021.

\bibitem{ghiasi2021SimpleCopyPaste}
G.~Ghiasi, Y.~Cui, A.~Srinivas, R.~Qian, T.-Y. Lin, E.~D. Cubuk, Q.~V. Le, and
  B.~Zoph, ``\BIBforeignlanguage{en}{Simple {Copy}-{Paste} {Is} a {Strong}
  {Data} {Augmentation} {Method} for {Instance} {Segmentation}},'' in
  \emph{\BIBforeignlanguage{en}{{CVPR}}}, 2021.

\bibitem{tang2021LookCloser}
C.~Tang, H.~Chen, X.~Li, J.~Li, Z.~Zhang, and X.~Hu,
  ``\BIBforeignlanguage{en}{Look {Closer} {To} {Segment} {Better}: {Boundary}
  {Patch} {Refinement} for {Instance} {Segmentation}},'' in
  \emph{\BIBforeignlanguage{en}{{CVPR}}}, 2021.

\bibitem{lai2020MASTMemoryAugmented}
Z.~Lai, E.~Lu, and W.~Xie, ``{MAST}: {A} {Memory}-{Augmented} {Self}-supervised
  {Tracker},'' in \emph{{CVPR}}, 2020.

\bibitem{garg2021MaskSelection}
S.~Garg and V.~Goel, ``\BIBforeignlanguage{en}{Mask {Selection} and
  {Propagation} for {Unsupervised} {Video} {Object} {Segmentation}},'' in
  \emph{\BIBforeignlanguage{en}{{WACV}}}, 2021.

\bibitem{wang2021DifferentTracking}
Z.~Wang, H.~Zhao, Y.-L. Li, S.~Wang, P.~H.~S. Torr, and L.~Bertinetto, ``Do
  {Different} {Tracking} {Tasks} {Require} {Different} {Appearance} {Models}?''
  in \emph{{NeurIPS}}, 2021.

\bibitem{russakovsky2015ImageNetLarge}
O.~Russakovsky, J.~Deng, H.~Su, J.~Krause, S.~Satheesh, S.~Ma, Z.~Huang,
  A.~Karpathy, A.~Khosla, M.~Bernstein, A.~C. Berg, and L.~Fei-Fei,
  ``\BIBforeignlanguage{en}{{ImageNet} {Large} {Scale} {Visual} {Recognition}
  {Challenge}},'' \emph{\BIBforeignlanguage{en}{International Journal of
  Computer Vision (IJCV)}}, vol. 115, no.~3, pp. 211--252, Dec. 2015.

\bibitem{he2020MomentumContrast}
K.~He, H.~Fan, Y.~Wu, S.~Xie, and R.~Girshick, ``Momentum {Contrast} for
  {Unsupervised} {Visual} {Representation} {Learning},'' in \emph{{CVPR}},
  2020.

\bibitem{jabri2020SpaceTimeCorrespondence}
A.~Jabri, A.~Owens, and A.~A. Efros, ``Space-{Time} {Correspondence} as a
  {Contrastive} {Random} {Walk},'' in \emph{{NeurIPS}}, 2020.

\bibitem{yan2022GrandUnification}
B.~Yan, Y.~Jiang, P.~Sun, D.~Wang, Z.~Yuan, P.~Luo, and H.~Lu, ``Towards
  {Grand} {Unification} of {Object} {Tracking},'' in \emph{{ECCV}}, 2022.

\bibitem{cai2022MeMOTMultiObject}
J.~Cai, M.~Xu, W.~Li, Y.~Xiong, W.~Xia, Z.~Tu, and S.~Soatto, ``{MeMOT}:
  {Multi}-{Object} {Tracking} with {Memory},'' in \emph{{CVPR}}, 2022.

\bibitem{korbarEndtoendTrackingMultiquery2022}
B.~Korbar and A.~Zisserman, ``End-to-end {Tracking} with a {Multi}-query
  {Transformer},'' Oct. 2022.

\bibitem{kuhn1955Hungarianmethod}
H.~W. Kuhn, ``The {Hungarian} method for the assignment problem,'' \emph{Naval
  Research Logistics Quarterly}, 1955.

\bibitem{luiten2021RobMOTSBenchmark}
J.~Luiten, A.~Hoffhues, B.~Beqa, P.~Voigtlaender, I.~Sárándi, P.~Dendorfer,
  A.~Osep, A.~Dave, T.~Khurana, T.~Fischer, X.~Li, Y.~Fan, P.~Tokmakov, S.~Bai,
  L.~Yang, F.~Perazzi, N.~Xu, A.~Bewley, J.~Valmadre, S.~Caelles,
  J.~Pont-Tuset, X.~Wang, A.~Geiger, F.~Yu, D.~Ramanan, L.~Leal-Taixé, and
  B.~Leibe, ``{RobMOTS} : {A} {Benchmark} and {Simple} {Baselines} for {Robust}
  {Multi}-{Object} {Tracking} and {Segmentation},'' in \emph{{CVPR} {RVSU}
  {Workshop}}, 2021.

\bibitem{he2017MaskRCNN}
K.~He, G.~Gkioxari, P.~Dollar, and R.~Girshick, ``Mask {R}-{CNN},'' in
  \emph{{ICCV}}, 2017.

\bibitem{luiten2018PReMVOSProposalGeneration}
J.~Luiten, P.~Voigtlaender, and B.~Leibe, ``{PReMVOS}: {Proposal}-{Generation},
  {Refinement} and {Merging} for {Video} {Object} {Segmentation},'' in
  \emph{{ACCV}}, 2018.

\bibitem{teed2020RAFTRecurrent}
Z.~Teed and J.~Deng, ``{RAFT}: {Recurrent} {All}-{Pairs} {Field} {Transforms}
  for {Optical} {Flow},'' in \emph{{ECCV}}, 2020.

\bibitem{pont-tuset20182017DAVIS}
J.~Pont-Tuset, F.~Perazzi, S.~Caelles, P.~Arbeláez, A.~Sorkine-Hornung, and
  L.~Van~Gool, ``The 2017 {DAVIS} {Challenge} on {Video} {Object}
  {Segmentation},'' \emph{arXiv:1704.00675}, Mar. 2018.

\bibitem{qiao2020ViPDeepLabLearning}
S.~Qiao, Y.~Zhu, H.~Adam, A.~Yuille, and L.-C. Chen, ``{ViP}-{DeepLab}:
  {Learning} {Visual} {Perception} with {Depth}-aware {Video} {Panoptic}
  {Segmentation},'' in \emph{{CVPR}}, 2021.

\bibitem{kim2020EagerMOTRealtime}
A.~Kim, A.~Ošep, and L.~Leal-Taixé, ``{EagerMOT}: {Real}-time {3D}
  multi-object tracking and segmentation via sensor fusion,'' in \emph{{CVPR} -
  {Workshops}}, 2020.

\bibitem{xu2020SegmentPoints}
Z.~Xu, W.~Zhang, X.~Tan, W.~Yang, H.~Huang, S.~Wen, E.~Ding, and L.~Huang,
  ``Segment as {Points} for {Efficient} {Online} {Multi}-{Object} {Tracking}
  and {Segmentation},'' in \emph{{ECCV}}, 2020.

\bibitem{song2020OnlineMultiObject}
Y.-m. Song and M.~Jeon, ``Online {Multi}-{Object} {Tracking} and {Segmentation}
  with {GMPHD} {Filter} and {Simple} {Affinity} {Fusion},'' in \emph{{CVPR} -
  {Workshops}}, 2020.

\bibitem{braso2022MultiObjectTracking}
G.~Braso, O.~Cetintas, and L.~Leal-Taixe, ``Multi-{Object} {Tracking} and
  {Segmentation} via {Neural} {Message} {Passing},'' Jul. 2022.

\bibitem{ahrnbom2021Realtimeonline}
M.~Ahrnbom, M.~Nilsson, and H.~Ardö, ``Real-time and online segmentation
  multi-target tracking with track revival re-identification,'' in
  \emph{International {Conference} on {Computer} {Vision} {Theory} and
  {Applications} ({VISAPP})}, 2021.

\bibitem{xing2009Multiobjecttracking}
J.~Xing, H.~Ai, and S.~Lao, ``Multi-object tracking through occlusions by local
  tracklets filtering and global tracklets association with detection
  responses,'' in \emph{{CVPR}}, 2009, pp. 1200--1207.

\bibitem{zhou2019OmniScaleFeature}
K.~Zhou, Y.~Yang, A.~Cavallaro, and T.~Xiang, ``Omni-{Scale} {Feature}
  {Learning} for {Person} {Re}-{Identification},'' in \emph{{ICCV}}, 2019.

\bibitem{miah2021EmpiricalAnalysis}
M.~Miah, J.~Pepin, N.~Saunier, and G.-A. Bilodeau, ``An {Empirical} {Analysis}
  of {Visual} {Features} for {Multiple} {Object} {Tracking} in {Urban}
  {Scenes},'' in \emph{International {Conference} on {Pattern} {Recognition}
  ({ICPR})}, 2021.

\bibitem{wojke2017Simpleonline}
N.~Wojke, A.~Bewley, and D.~Paulus, ``Simple online and realtime tracking with
  a deep association metric,'' in \emph{International {Conference} on {Image}
  {Processing} ({ICIP})}, 2017.

\bibitem{ristani2018FeaturesMultiTarget}
E.~Ristani and C.~Tomasi, ``Features for {Multi}-{Target} {Multi}-{Camera}
  {Tracking} and {Re}-{Identification},'' in \emph{{CVPR}}, 2018.

\end{thebibliography}

\end{document}